\documentclass[runningheads]{llncs}

\usepackage{eccv}
\usepackage{eccvabbrv}

\usepackage{graphicx}
\usepackage{booktabs}
\usepackage{float}

\usepackage[accsupp]{axessibility}
\usepackage{hyperref}

\usepackage{orcidlink}
\usepackage[subpreambles=true]{standalone}

\begin{document}

\title{PixGS: Pixel-Space Diffusion for Direct 3D Gaussian Splat Generation} 

\author{Cao Duy \qquad Phong Nguyen}
\authorrunning{C. Duy and P. Nguyen}
\renewcommand{\thefootnote}{\fnsymbol{footnote}}
\institute{
Qualcomm AI Research\footnotemark[1]\\
\texttt{\{duycao, phongnh\}@qti.qualcomm.com}
}
\footnotetext[1]{Qualcomm AI Research is an initiative of Qualcomm Technologies, Inc.}
\renewcommand{\thefootnote}{\arabic{footnote}}

\maketitle

\begin{abstract}
Recent advances in 3D content generation from text or images have achieved impressive results, yet view inconsistency from 2D generators and the scarcity of high-quality 3D data remain significant bottlenecks. Existing solutions~\cite{lin2025diffsplatrepurposingimagediffusion, meng2025zero1togtamingpretrained2d} typically adapt large-scale pre-trained text-to-image latent diffusion models to generate 3D Gaussian Splats (3DGS). However, these approaches often rely on training complex cascade pipelines that are computationally expensive and scalability-limited. Most critically, the quality of generated 3D assets is inherently constrained by each component capacity and compressed latent space, leading to decoding artifacts and accumulated errors. To address these limitations, we propose \textbf{PixGS}, a single-stage pipeline for direct high-quality 3DGS generation, which leverages recent advances in pixel-space diffusion to bypass lossy latent compression while still benefiting from the vast 2D generative priors. By directly denoising 3D Gaussian attributes at each timestep, our method enables precise, splat-level regularization of both appearance and geometry. Furthermore, we introduce a comprehensive supervision strategy that incorporates surface normals, depth, and high-frequency structural information, which is often overlooked in prior works. Experiments demonstrate that PixGS outperforms current state-of-the-art methods while maintaining a fast inference speed ($\approx 1s$ on a single A100 GPU), offering a robust and efficient alternative to multi-stage generation pipelines.
\end{abstract}

\section{Introduction}
\label{sec:intro}
While recent 3D generative advances often favor mesh-based representations~\cite{zhang2024clay, lai2025hunyuan3d25highfidelity3d, hong20243dtopia, lan2024ln3diff} for their geometric clarity in animation and printing, domains such as VR/AR, gaming and film prioritize visual realism and rendering efficiency. This has cemented 3D Gaussian Splats (3DGS)~\cite{kerbl3Dgaussians,kheradmand20243d} as a pivotal representation, fueling research efforts to integrate its superior rendering capabilities into robust generative pipelines.

Several existing methods~\cite{tang2024lgm, xu2024instantmesh, wang2024crm} attempt to synthesize 3DGS using multi-view-aware 2D diffusion models~\cite{shi2023MVDream, liu2023syncdreamer} to generate consistent images of an object, followed by a separate 3D reconstruction stage. While promising, the final 3D quality is strictly capped by the upstream 2D generator, and any slight inconsistency in the generated views leads to artifacts (floaters) during reconstruction.

A recent promising direction involves directly generating 3DGS in an image-like format. Inspired by generalizable 3DGS reconstruction techniques~\cite{szymanowicz24splatter, charatan2024pixelsplat3dgaussiansplats}, DiffSplat~\cite{lin2025diffsplatrepurposingimagediffusion} finetunes latent diffusion models to generate 2D tensors where each pixel encodes the attributes of a 3D Gaussian. While this provides multi-view consistency and leverages 2D priors, it introduces a significant architectural bottleneck. The pipeline requires training three interdependent models: a 3DGS reconstructor, a VAE to compress attributes into a latent space, and a latent diffusion model. This cascaded training regime is computationally expensive and introduces substantial engineering overhead, as each stage must converge reasonably well before the next can proceed, making the overall system fragile and difficult to scale. Also, to mitigate the computation costs of repeated Gaussian decoding during diffusion training, DiffSplat employs an additional lightweight decoder, trained from scratch, which reduces memory usage at the expense of increased system complexity. Moreover, reliance on a compressed latent space inherently limits the representational capacity and fine-grained detail of the generated 3DGS. 

We propose PixGS, a single-stage pipeline for high-quality 3DGS generation, designed to address the limitations of previous works. PixGS leverages advances in large-scale Pixel-space Image Diffusion~\cite{chen2025pixelflow,yu2025pixeldit,wang2025pixnerdpixelneuralfield,ma2025decofrequencydecoupledpixeldiffusion, ma2026pixelgenpixeldiffusionbeats}  models trained on extensive image corpora. To capitalize on the powerful 3D geometry estimation priors inherent in web-scale image diffusion models demonstrated in prior research~\cite{lin2025diffsplatrepurposingimagediffusion, sweetdreamer, fu2024geowizardunleashingdiffusionpriors, ke2024repurposingdiffusionbasedimagegenerators}, we represent 3D objects using Gaussian attribute tensors, which are 2D grids where each pixel encodes the geometric and textural properties of the 3DGS. Instead of relying on 3D latent space, our generative model learns and operates directly at the splat distribution, denoising and regularizing Gaussians at each timestep. To leverage this, we introduce a comprehensive loss set, including depth and normal supervision, as well as Laplacian of Gaussian~\cite{marr1980theory} (LoG) loss for high-frequency feature regularization. By bypassing the training requirements of VAE latent compression and integrating LoG loss, we successfully capture intricate details that previous methods often fail to recover (such as text and thin structures) while maintaining computational efficiency. Alongside the applied loss functions, we introduce a three-phase training schedule that decouples the generative model from the constraints of the reconstruction model. Our contributions can be summarized as follows:
\begin{itemize}
    \item We introduce PixGS, a single-stage pipeline for high-quality 3DGS generation that bypasses the limitations of latent compression and is highly scalable.
    \item We adapt 2D pixel-space diffusion to operate on Gaussian attribute tensors while preserving powerful pre-trained 2D priors to ensure rapid convergence.
    \item We integrate a suite of loss functions designed for the direct supervision of 3DGS geometry and appearance, focusing on high-frequency details and a three-phase training scheme that progressively enhances generation quality beyond the capabilities of the initial 3DGS reconstructor.
    \item Extensive experiments demonstrating the strong performance of PixGS, supported by ablation studies that validate the effectiveness of each design choice.
\end{itemize}

\section{Related Work}
\noindent\textbf{Optimization-based generative models.} Early breakthroughs in 3D generation were dominated by optimization-based methods~\cite{poole2022dreamfusion, lin2023magic3d}. These approaches leverage the visual richness of 2D diffusion models to iteratively refine 3D representations, typically NeRFs, through Score Distillation Sampling (SDS), ensuring their rendered views align with a text prompt. Following the introduction of 3D Gaussian Splatting, frameworks~\cite{tang2023dreamgaussian, chen2023gsgen, yi2024gaussiandreamerfastgenerationtext} adapted these optimization strategies to 3DGS representation, significantly accelerating the synthesis process. While these methods require no ground-truth 3D data and inherit the creative capacity of 2D foundations, they are notoriously slow due to their per-scene optimization requirement. Furthermore, they frequently suffer from over-smoothed geometry or multi-view inconsistencies, such as ``Janus" problem~\cite{wang2023prolificdreamer}.

\noindent\textbf{3D-native generative models} To facilitate feed-forward synthesis, 3D-native generative models~\cite{nichol2022pointe, jun2023shape, zhang2024gaussiancube, zhao2023michelangelo} have emerged to learn 3D shape distributions directly from geometric data. These approaches achieve rapid inference and maintain superior geometric consistency compared to optimization-based methods. However, they are fundamentally constrained by the scale of available 3D datasets~\cite{objaverse, objaverseXL}, which are orders of magnitude smaller than web-scale image corpora. Consequently, capturing high-quality textures and complex topologies often necessitates massive architectural capacity, as seen in recent frameworks~\cite{zhang2024clay, lai2025hunyuan3d25highfidelity3d, xiang2024structured, xiang2025nativecompactstructuredlatents}. Despite their scale, these models often struggle to generalize beyond their specific training distributions compared to methods that leverage 2D foundation priors.

\noindent\textbf{Reconstruction-based 3D Generative Models} A prominent direction involves 2D prior utilization through a multi-stage process: first producing multi-view-consistent 2D images~\cite{shi2023MVDream, liu2023syncdreamer, tang2023mvdiffusionenablingholisticmultiview} and then mapping them to 3D space via a separate reconstructor~\cite{tang2024lgm, xu2024instantmesh, wang2024crm, xu2024grm, chen2024laraefficientlargebaselineradiance}. However, this sequential architecture introduces a significant geometric bottleneck. Because the reconstructor treats the generated images as fixed ground truth, any subtle inconsistencies in perspective or texture between views lead to accumulated errors and ``floaters" highlighting the inherent difficulty of maintaining geometric integrity in decoupled pipelines.

\noindent\textbf{Structured 3DGS Generation} Building on generalizable reconstruction frameworks~\cite{szymanowicz24splatter, charatan2024pixelsplat3dgaussiansplats, yu2024gsdf}, recent methods~\cite{lin2025diffsplatrepurposingimagediffusion, meng2025zero1togtamingpretrained2d} attempt to directly synthesize 3DGS as structured 2D attribute tensors. While this image-like format facilitates multi-view consistency, it typically necessitates a complex, multi-stage pipeline. The cascaded regime introduces substantial engineering overhead and training fragility, as it requires the sequential convergence of interdependent components. Conversely, single-stage approaches like DiffusionGS~\cite{diffusiongs} attempt to generate 3DGS directly through a diffusion process. However, these models typically rely on specialized architectures trained entirely from scratch on synthetic 3D datasets. By discarding pre-trained 2D priors, these models suffer from slow convergence and limited generalization, as they must learn basic visual semantics and appearance cues solely from a relatively small 3D training corpus. Our method addresses the limitations of previous works by introducing a single-stage framework leveraging large-scale 2D priors, with fast inference that produces high-quality 3D content.

\section{Background}
\subsection{Pixel-Space Diffusion}
Pixel-space diffusion models~\cite{chen2025pixelflow,yu2025pixeldit,wang2025pixnerdpixelneuralfield,ma2025decofrequencydecoupledpixeldiffusion, ma2026pixelgenpixeldiffusionbeats} generate images by learning a velocity field $v_\theta(\mathbf{x}_t, t)$ on the raw pixel manifold $\mathbf{x} \in \mathbb{R}^{H \times W \times 3}$. Following the Flow Matching formulation~\cite{lipman2023flowmatchinggenerativemodeling}, the forward process defines a probability path $\mathbf{x}_t = t\mathbf{x}_1 + (1-t)\mathbf{x}_0$ between data $\mathbf{x}_1$ and noise $\mathbf{x}_0 \sim \mathcal{N}(0, \mathbf{I})$, with $t \in [0,1]$, where the model is trained to predict the constant velocity $\mathbf{v}_t = \mathbf{x}_1 - \mathbf{x}_0$. To maintain computational efficiency without a VAE, these models partition the image into large patches $\mathbf{X} = \{\mathbf{X}^n\}_{n=1}^N$ (e.g., $16 \times 16$).

However, standard linear decoders struggle to recover high-frequency details from large, compressed tokens. PixNerd~\cite{wang2025pixnerdpixelneuralfield} improves this by introducing a patch-wise neural field. For each patch token $\mathbf{X}^n$, the transformer predicts the weights $\mathcal{W}^n$ of a local MLP: $\mathcal{W}^n = \text{Linear}(\text{SiLU}(\mathbf{X}^n))$. The velocity $\mathbf{v}^n(i, j)$ for a specific pixel coordinate $(i, j)$ is then decoded as:
\begin{equation}
    \mathbf{v}^n(i, j) = \text{MLP}\left( \text{Concat}[\text{PE}(i, j), \mathbf{x}_t^n(i, j)] \mid \mathcal{W}^n \right),
\end{equation}
where $\text{PE}$ is a coordinate encoding (DCT-basis~\cite{ahmed1974dct}). By replacing fixed linear projections with coordinate-conditioned neural fields, PixNerd achieves high-fidelity, single-stage generation directly in pixel space.

\subsection{Splatter Image}
Splatter Image~\cite{szymanowicz24splatter} represents a 3D scene as an image-structured collection of 3D Gaussians. Unlike unordered point clouds, it utilizes a 2D image of size $H \times W$ as a container, where each pixel $u$ stores the parameters of exactly one 3D Gaussian, where dimension $K$ include the Gaussian properties: opacity $\alpha$, color $c$, covariance $\Sigma$, and center $\mu$. Subsequently, an image-to-image network is employed to map the input image into a gaussian tensor $M \in \mathbb{R}^{K \times H \times W}$.

To reconstruct the 3D geometry from a single view, the center $\mu$ of a Gaussian at pixel $u = (u_1, u_2)$ is parameterized by a predicted depth $d$ and a 3D offset $\Delta = (\Delta_x, \Delta_y, \Delta_z)$:
\begin{equation}
    \mu = \begin{bmatrix} u_1 d + \Delta_x \\ u_2 d + \Delta_y \\ d + \Delta_z \end{bmatrix}.
\end{equation}
While the Gaussians are initially aligned with camera rays, the predicted 3D offsets $\Delta$ allow the network to ``splat" Gaussians into unobserved regions, enabling a full $360^\circ$ reconstruction from a single viewpoint.

\subsection{3DGS Rendering}
3D Gaussian Splatting (3DGS)~\cite{kerbl3Dgaussians, kheradmand20243d} represents a 3D scene using a set of explicit, translucent Gaussian primitives. Each Gaussian is parameterized by its center $\mu \in \mathbb{R}^3$, opacity $\alpha \in [0, 1]$, color $c$ (often represented via Spherical Harmonics), and a 3D covariance matrix $\Sigma$. To ensure physical validity during optimization, $\Sigma$ is factored into a scaling matrix $S$ and a rotation matrix $R$ as $\Sigma = RSS^\top R^\top$. 

To render the scene from a given viewpoint, 3D Gaussians are projected into 2D image space. Using a local affine approximation of the camera transformation, the projected 2D covariance $\Sigma'$ is computed as:
\begin{equation}
    \Sigma' = JW\Sigma W^\top J^\top,
\end{equation}
where $W$ is the world-to-camera transformation matrix and $J$ is the Jacobian of the projective transformation. The final color $C$ for a pixel is determined by sorting the $N$ overlapping Gaussians by depth and performing differentiable $\alpha$-blending:
\begin{equation}
    C = \sum_{i=1}^{N} c_i \sigma_i \prod_{j=1}^{i-1} (1 - \sigma_j),
\end{equation}
where $\sigma_i$ is the density of the $i$-th Gaussian at that pixel, calculated by evaluating the 2D Gaussian distribution multiplied by the opacity $\alpha_i$.

\section{Methodology}
\begin{figure}[t!]
    \centering
    \includegraphics[width=0.9\textwidth]{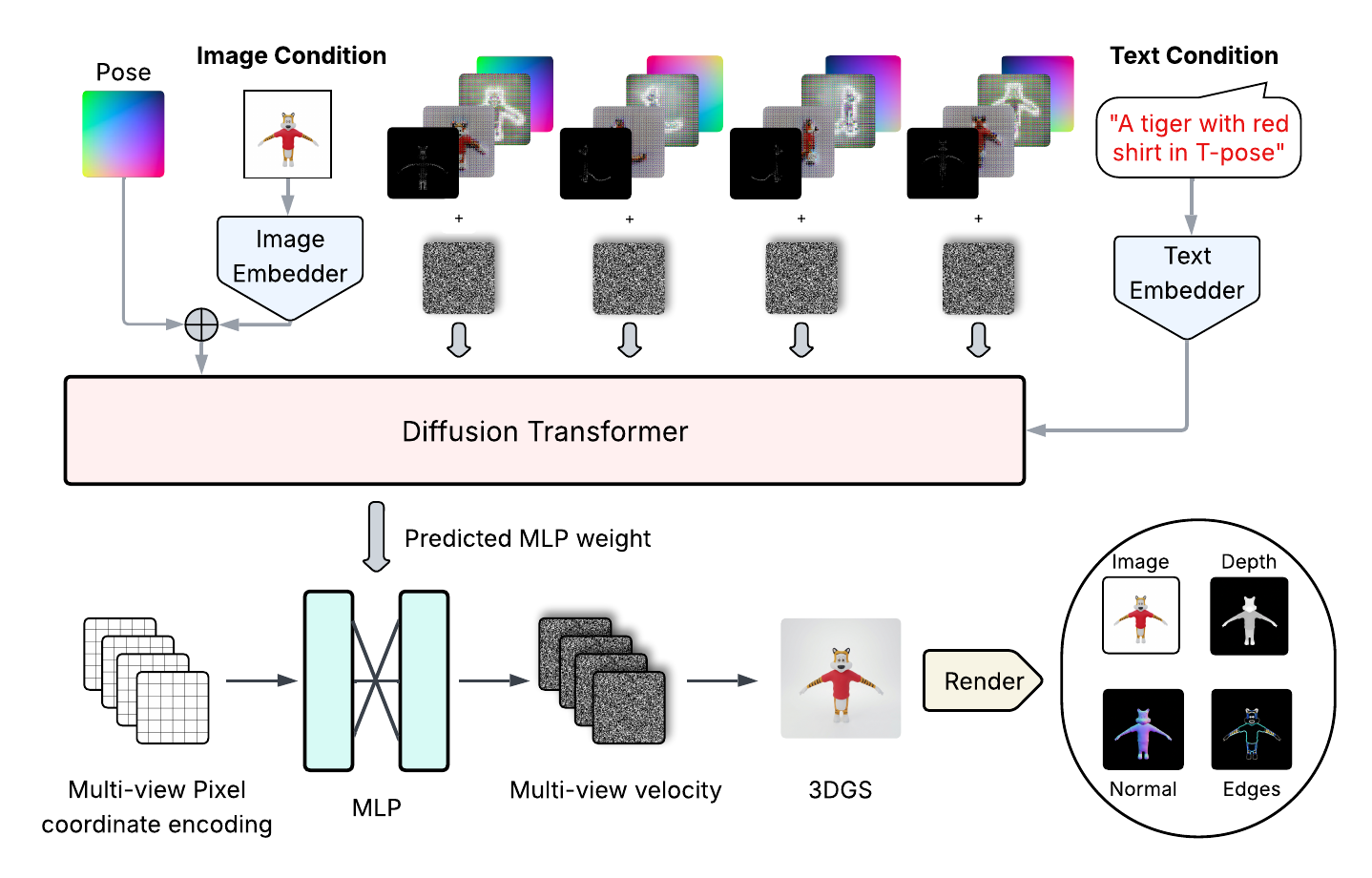}
    \caption{\textbf{Pipeline Overview}. PixGS directly denoises 3D Gaussian attribute tensors conditioned on image and text prompts utilizing 2D priors from Pixel Diffusion models. $\oplus$ denotes the concatenation of features.}
    \label{fig:pipeline}
\end{figure}
PixGS is a single-stage, pixel-space diffusion model that directly generates 3DGS parameters bypassing the limitations of a latent bottleneck. Our motivation for adopting an image-like 3D representation (Sec.~\ref{representation}) is to leverage established 2D image priors while avoiding the information loss associated with VAE compression. We introduce a suite of tailored design choices (Sec.~\ref{pipeline}), objectives (Sec.~\ref{sec:objectives}), and training schemes (Sec.~\ref{sec:scheme}) that adapt pixel-level diffusion to the requirements of 3DGS generation.

\subsection{3D Representation}\label{representation}
Following DiffSplat~\cite{lin2025diffsplatrepurposingimagediffusion}, we parameterize our 3D representation as a grid of pixel-aligned Gaussians, where each primitive $\mathbf{g}_i \in \mathbb{R}^{12}$ consists of color $\mathbf{c}$, scale $\mathbf{s}$, rotation $\mathbf{r}$, opacity $o$, and depth $d$. The 3D position $\mathbf{x}$ can be unprojected from the 2D image plane using the camera's intrinsic $\mathbf{K}$ and extrinsic $\{\mathbf{R}, \mathbf{t}\}$ matrices: $\mathbf{x} = \mathbf{R}^\top \mathbf{K}^{-1} [\mathbf{u} | d] - \mathbf{t}$, with $[\mathbf{u} | d]$ is homogeneous pixel coordinates $\mathbf{u}\in \mathbb{R}^{2}$ with depth. 

We represent each 3D object as $\mathcal{G} \in \mathbb{R}^{V_{in} \times g \times H \times W}$ tensor, where $H \times W$ denotes the spatial resolution of a 2D grid and each pixel encodes the attributes of a single Gaussian primitive and $g$ denotes the number of Gaussian parameters. Here, $V_{in}$ represents the number of grids, each corresponding to a camera viewpoint that spatially covers the object, resulting in a total of $V_{in} \times H \times W$ Gaussians. Intuitively, this representation is analogous to a multi-view image set where Gaussian attributes replace standard RGB channels, thereby facilitating the seamless adaptation of 2D pixel-space diffusion models to 3D generation. To bootstrap the training process, we obtain pseudo-labels from off-the-shelf 3DGS reconstructors that follow a Splatter Image-style~\cite{szymanowicz24splatter} representation, mapping multi-view images into our target attribute space. Specifically, we utilize the reconstruction capability of GSRecon~\cite{lin2025diffsplatrepurposingimagediffusion}, fine-tuning it to generate pseudo-labels for training our generative model.

\subsection{Generative model}\label{pipeline}
We adapt PixNerd~\cite{wang2025pixnerdpixelneuralfield} architecture to the 3D domain, leveraging its extensive pre-trained priors to facilitate Gaussian generation. By building upon this framework, our model inherits rich 2D knowledge acquired from a large-scale corpus of approximately 45 million images, which provides a robust foundation for capturing complex visual details.

\subsubsection{Model Architecture}
An overview of our proposed PixGS pipeline is illustrated in Fig.~\ref{fig:pipeline}. Given a batch size $B$ of Gaussian attribute tensors $\mathcal{G}_1 \in \mathbb{R}^{B \times V_{in} \times g \times H \times W}$, we define a linear probability path following Flow Matching formulation~\cite{lipman2023flowmatchinggenerativemodeling}. The forward diffusion process interpolates between the data tensor $\mathcal{G}_1$ and a Gaussian noise tensor $\mathcal{G}_0 \sim \mathcal{N}(0, \mathbf{I})$ as follows:
\begin{equation}
    \mathcal{G}_t = t\mathcal{G}_1 + (1-t)\mathcal{G}_0, \quad t \in [0, 1].
\end{equation}
To provide the model with geometric guidance, we augment the noisy tensor with Plücker coordinates~\cite{sitzmann2021lfns} $\mathcal{P} \in \mathbb{R}^{V_{in} \times d_{pl} \times H \times W}$ representing the camera rays for each viewpoint. The resulting augmented tensor is denoted as $\mathcal{G}'_t \in \mathbb{R}^{B \times V_{in} \times c \times H \times W}$, where $c = g + d_{pl}$ is the total channel dimension.

Following the pixel-space diffusion convention, we partition the spatial dimensions of each viewpoint into large non-overlapping patches of size $P \times P$ (e.g. $P=16$). These patches are flattened and concatenated to form a sequence of tokens $\mathbf{X} \in \mathbb{R}^{(B \cdot V_{in}) \times L \times D}$, where $L = \frac{HW}{P^2}$ is the sequence length per view and $D = P^2 \cdot c$ is the token dimensionality. 

To ensure cross-view geometric consistency, we extend the standard Attention block to a Multi-view Attention block~\cite{huang2024mvadapter}. By reshaping the token sequence to $\mathbf{X} \in \mathbb{R}^{B \times (V_{in} \cdot L) \times D}$ before attention computation, each patch is allowed to attend to all other patches across all viewpoints of the same 3D object. We also adapt RoPE2d~\cite{su2023roformerenhancedtransformerrotary} in Pixel Diffusion into a Multi-view RoPE2d. Rather than applying embeddings independently to each view, we treat $V_{in}$ viewpoints as spatially aligned segments of a high-resolution composite image. Specifically, we scale the RoPE2d coordinate manifold to a virtual resolution of $(\frac{V_{in}}{2} H) \times (\frac{V_{in}}{2} W)$ to accommodate the collective resolution of all views, assigning unique positional embeddings across $V_{in} \cdot L$ tokens. This combination allows the transformer to learn relative spatial relationships not only within a single view but across the entire multi-view collective.

Following the transformer blocks, the final hidden states $\mathbf{X}^n_v$, representing the $n$-th patch of viewpoint $v$-th, are used to parameterize a local neural field. We predict the weights $\mathcal{W}^n_v$ of a patch-specific MLP via a linear projection of the conditioned hidden states:
\begin{equation}
    \mathcal{W}^n_v = \text{Linear}(\text{SiLU}(\mathbf{X}^n_v)).
\end{equation}
To denoise Gaussian splats directly in pixel space, the velocity $\mathbf{v}$ at a specific multi-view pixel coordinate $(v, i, j)$ within the patch is decoded as:
\begin{equation}
    \mathbf{v}^n(v, i, j) = \text{MLP}\left(\text{Concat}[\text{PE}(v, i, j), \mathcal{G'}_t^n(v, i, j)] \mid \mathcal{W}^n_v \right),
\end{equation}
where $\text{PE}$ denotes a coordinate encoding and $\mathcal{G'}_t^n$ is the noisy input at patch $n$.

\subsubsection{Image-conditioned Paradigms}
\begin{figure}[t]
    \centering
    \begin{subfigure}{0.46\textwidth}
        \centering
        \includegraphics[width=\linewidth]{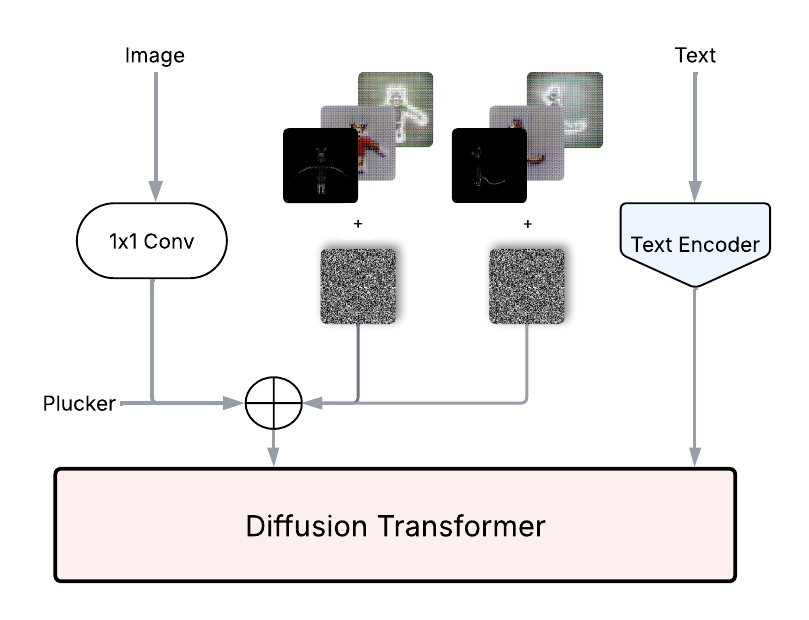}
        \caption{Viewpoint Concatenation}
        \label{fig:viewconcat}
    \end{subfigure}
    \hfill
    \begin{subfigure}{0.53\textwidth}
        \centering
        \includegraphics[width=\linewidth]{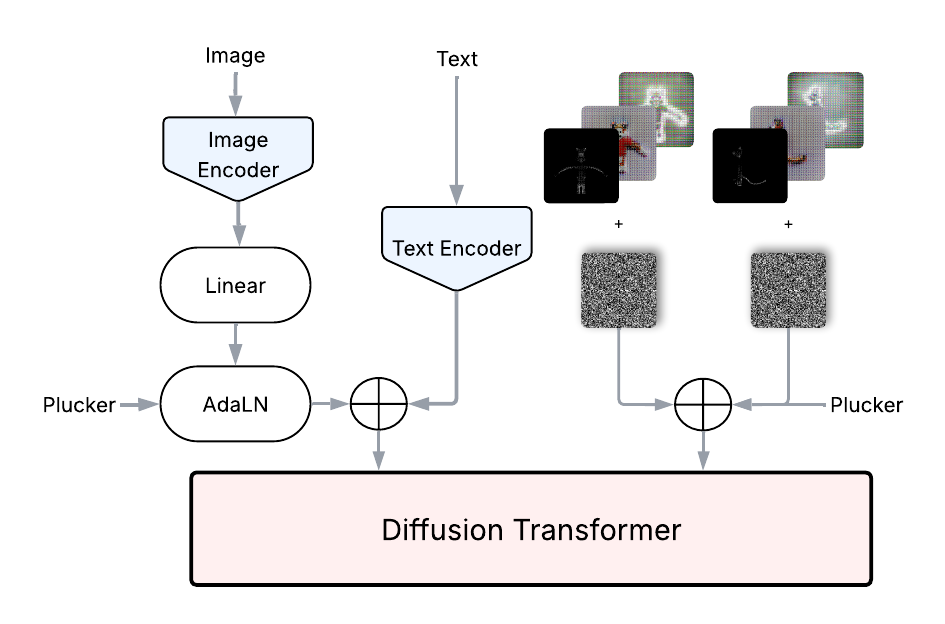}
        \caption{Image Prompt Adapter}
        \label{fig:ipadapter}
    \end{subfigure}
    
    \caption{\textbf{Paradigms for image-conditioned generation.} We explore two strategies: (a) The reference image is integrated as an additional viewpoint (b) Semantic features are extracted and injected via cross-attention. $\oplus$ denotes the concatenation of features.}
    \label{fig:conditioning_paradigms}
\end{figure}
Our objective is to develop a 3D generative framework that can be conditioned on either text or images. However, since text-to-image pixel diffusion models do not natively support visual prompts, we must adapt the architecture to accommodate the image modality. To this end, we explore two widely adopted approaches for image-conditioned generation: View Concatenation~\cite{diffusiongs, lin2025diffsplatrepurposingimagediffusion} and Image Prompt Adapter~\cite{ye2023ipadaptertextcompatibleimage, xiang2024structured}. We provide a schematic overview of these approaches in Fig.~\ref{fig:conditioning_paradigms} and a comparison of their performance in our ablation study (Sec.~\ref{sec:ablation_imagecond}).

\paragraph{Viewpoint Concatenation:} In this configuration, we incorporate the reference image $I_{cond}$ as a guided viewpoint by projecting its RGB channels to match the Gaussian attribute dimension $g$ and append the Plücker coordinates representing the reference camera pose. This conditioned feature map is then concatenated along the view dimension with the noisy attribute tensor $\mathcal{G}_t$.

\paragraph{Image Prompt Adapter:} This approach leverages DINO-v2~\cite{oquab2023dinov2} to extract features from $I_{cond}$. We employ an AdaLN~\cite{peebles2023scalable} module to inject viewpoint information, where embedding is scaled and shifted based on the reference Plücker coordinates. The image-embedding is subsequently projected to match the text-embedding dimensionality and are prepended to the textual prompt tokens.

\subsection{Training Objectives}\label{sec:objectives}
Our model is trained using a multi-task objective that combines generative modeling with pixel-aligned rendering supervision and structural regularization.

\paragraph{Flow Matching Loss:} Following Flow Matching~\cite{lipman2023flowmatchinggenerativemodeling} formulation, the model $v_\theta$ is trained to predict the constant velocity $v_t = \mathcal{G}_1 - \mathcal{G}_0$ that moves noise toward the data manifold:
\begin{equation}
    \mathcal{L}_{\text{FM}} = \mathbb{E}_{t, \mathcal{G}_0, \mathcal{G}_1} \left[ \| v_\theta(\mathcal{G}_t, t) - (\mathcal{G}_1 - \mathcal{G}_0) \|^2_2 \right],
\end{equation}
where $t$ is uniformly sampled from $[0, 1]$. As analyzed in Sec.~\ref{sec:ablation_loss}, relying exclusively on flow matching supervision results in poor 3D consistency, as the model fits the pseudo-label distribution without understanding underlying geometric constraints. To ensure physical plausibility, we complement the flow matching objective with direct supervision of 3D appearance and geometry.

\paragraph{Appearance Loss:} Although our model produces a large number of Gaussians, many initially appear as low-opacity background artifacts (Fig.~\ref{fig:only_fm}). To maximize primitive utilization, we perform rendering at twice the spatial resolution of the attribute tensor $\mathcal{G}_{1}$ ($2H \times 2W$). This super-resolution supervision provides a dense gradient signal, forcing the model to concentrate Gaussian density on valid object surfaces and recover high-frequency details that exceed the base grid resolution.
\begin{equation}
    \mathcal{L}_{\text{app}}= \frac{1}{V}\sum_{v=1}^{V}\ \left(\mathcal{L}_{\text{MSE}}(I_v,I^{\text{GT}}_v)+\lambda_p \mathcal{L}_{\text{LPIPS}}(I_v,I^{\text{GT}}_v)+\mathcal{L}_{\text{MSE}}(M_v,M^{\text{GT}}_v)\right),
\end{equation}
where $I_v$ and $M_v$ are the RGB images and silhouette masks differentiably rendered from the predicted 3D Gaussian tensor $\mathcal{G}_{1}$ via 3DGS rasterization~\cite{zhang2024radegsrasterizingdepthgaussian}, and $V$ denotes the number of rendered supervision views.

\paragraph{Geometry Loss:} We regularize the underlying geometry using surface normal and depth supervision:
\begin{equation}
    \mathcal{L}_{\text{geo}} = \frac{1}{V}\sum_{v=1}^{V}\
    \left(\lambda_d \| D_v - D_v^{GT} \|_2 + \lambda_n (1 - \cos(N_v, N_v^{GT}))\right),
\end{equation}
where $D_v$ and $N_v$ are the rendered depth and surface normal maps, respectively. 

\paragraph{High-frequency Loss:} To capture actively intricate details, we introduce a Multi-Scale Laplacian of Gaussian (LoG) loss $\mathcal{L}_{\text{LoG}}$. More detailed analysis on $\mathcal{L}_{\text{LoG}}$ will be provided in the Appendix. The LoG loss operates on the rendered images to emphasize high-frequency feature alignment:
\begin{equation}
    \mathcal{L}_{\text{LoG}} =  \frac{1}{V}\sum_{v=1}^{V}\
     \left(\sum_{\sigma \in \mathcal{S}} \omega_\sigma \| \Delta(G_\sigma * I_v) - \Delta(G_\sigma * I_v^{GT}) \|_1\right),
\end{equation}

where $\mathcal{S}$ denotes the set of standard deviations for the Gaussian kernels, $\Delta$ is the Laplace operator and $G_\sigma$ is a Gaussian kernel, and scale-specific weights $\omega_\sigma \propto \sigma^2$, balances the contribution of sharp features against rendering noise. The total objective is defined as:
\begin{equation}
    \mathcal{L}_{\text{total}} = \lambda_{f}\mathcal{L}_{\text{FM}} + \lambda_{a}\mathcal{L}_{\text{app}} + \lambda_{g}\mathcal{L}_{\text{geo}} +
    \lambda_{l}\mathcal{L}_{\text{LoG}}.
\end{equation}

\subsection{Training Scheme}\label{sec:scheme}
We utilize GSRecon solely for early guidance through a three-phase training schedule: (1) flow matching on pseudo-labels, (2) flow matching combined with ground-truth (GT) rendering supervision defined in Sec.~\ref{sec:objectives}, and (3) GT-only supervision. Under this schedule, pseudo-labels are mainly used to bootstrap convergence and improve training stability, rather than to determine PixGS’s final quality. This is because they are limited to the first 50K iterations, whereas the longest stage, phase~3, relies entirely on GT-only supervision for 150K iterations. Phase~2 then serves as a smooth transition from pseudo-label pretraining to GT-only fine-tuning, improving robustness with limited overhead.

Because 3DGS~\cite{kerbl3Dgaussians, kheradmand20243d} imposes strict requirements on attribute value ranges and properties: coordinates, color, opacity, and scale must remain within strict range, while rotations must maintain unit norm. Applying rendering losses prematurely when there is significant noise in early iterations causes numerical instability, leading to NaN gradients. Consequently, we use the flow matching loss with guidance from pseudo-labels in early stages to gradually map the model output to 3DGS attribute distribution, facilitating a stable transition to Phase 2 for appearance and geometry supervision. At this stage, because background Gaussians have been suppressed to near-zero opacity, the rendering process is significantly faster, leading to shorter iteration times.

However, the diffusion objective alone cannot perfectly learn the valid 3DGS attribute format. While we $L_2$-normalize rotations and clamp other attributes during the forward pass, these operations can introduce discrepancies that make rendering losses unreliable. We observe that although diffusion training eventually produces values close to the correct range, the unit-norm rotation property is particularly difficult to learn solely through flow matching. To address this, we introduce an explicit regularization term for the rotation attribute norm:
\begin{equation}
    \mathcal{L}_{\text{rot}} = \mathbb{E} \left[ (\|\mathbf{r}\|_2 - 1)^2 \right],
\end{equation}
where $\mathbf{r}$ denotes the predicted rotation quaternion. This training scheme ensures numerical stability and facilitates the rapid convergence of our generative model.

\section{Experiments}
\subsection{Experimental Settings}
\paragraph{Datasets:} We train PixGS on G-Objaverse and G-Objaverse-XL Alignment~\cite{qiu2023richdreamer}, which are curated high-quality subsets derived from Objaverse~\cite{objaverse} and Objaverse-XL~\cite{objaverseXL}, respectively. We further refined these collections by filtering assets based on aesthetic scores, resulting in a final training corpus of over 500K 3D objects. Each asset is rendered from 38 diverse viewpoints to ensure comprehensive spatial coverage, with corresponding textual descriptions provided by Cap3D~\cite{luo2023scalable}.

\paragraph{Text-Conditioned:} To evaluate the quality of our text-to-3D synthesis, we utilize 300 prompts from T3Bench~\cite{he2023t3bench} benchmark. These prompts are categorized into three levels of complexity: single object, objects with surroundings, and multi objects. We quantify the alignment between the input text and the rendered 3D results using CLIP similarity~\cite{radford2021learningtransferablevisualmodels} and CLIP R-Precision~\cite{park2021benchmark} (ViT-B/32). Furthermore, we employ ImageReward~\cite{xu2023imagerewardlearningevaluatinghuman} to assess the generated assets according to human aesthetic preferences.

\paragraph{Image-conditioned:} For tasks involving image-conditioned generation and reconstruction, we benchmark our model on 300 randomly selected objects from GSO~\cite{downs2022google} dataset, which remain unseen during training. We evaluate the quality of the synthesized views against GT renders using standard reconstruction metrics, including PSNR, SSIM, and LPIPS~\cite{zhang2018unreasonable}, all quantitative results are computed and averaged across multiple sampled viewpoints. 

\subsection{Text-conditioned Generation}
\begin{figure*}[t]
    \centering
    \includegraphics[width=1.\textwidth]{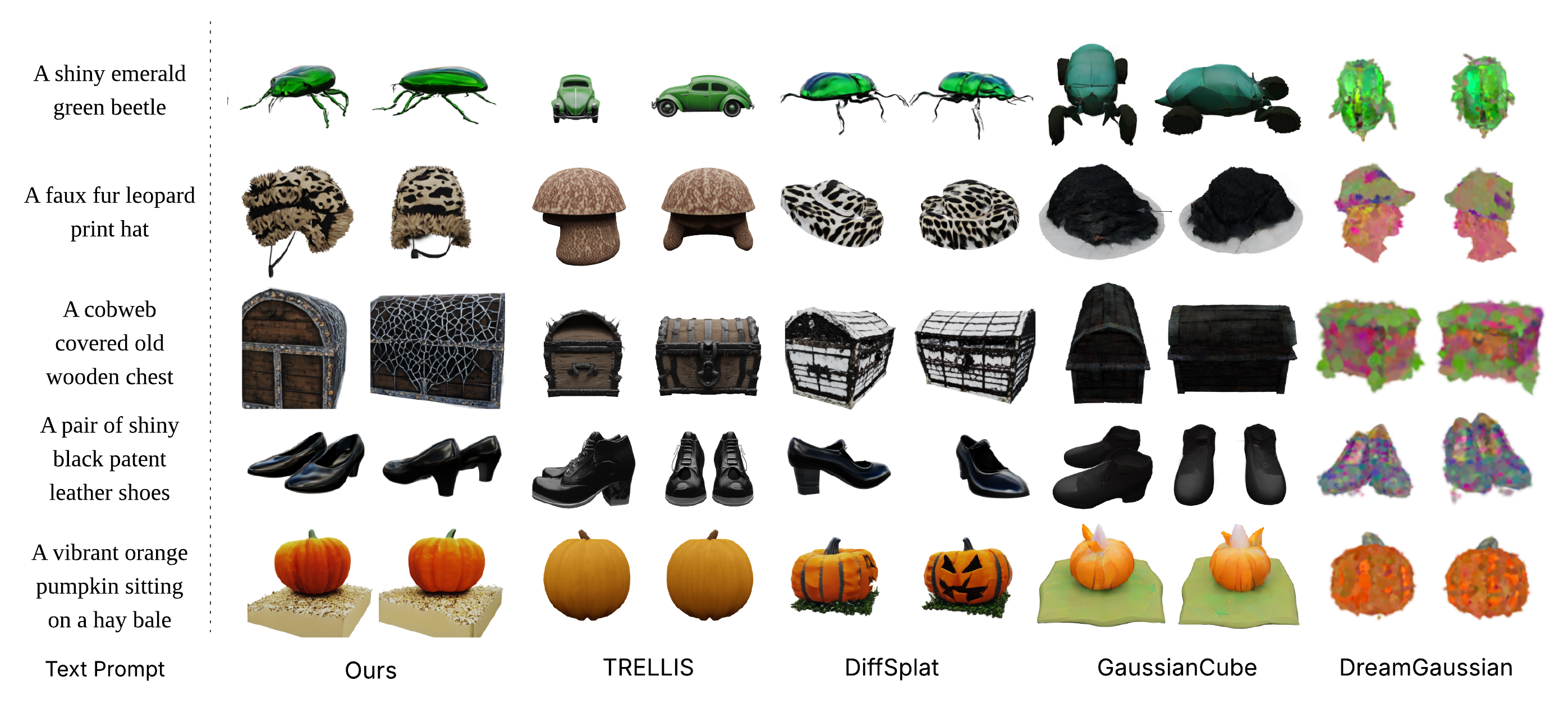}
    \caption{\textbf{Qualitative Results and Comparisons on Text-conditioned 3D Generation.} Best viewed \textbf{ZOOMED-IN}.}
    \label{fig:text_qualitative}
\end{figure*}

\paragraph{Baseline:} We benchmark our method against 5 state-of-the-art baselines capable of generating 3DGS representations. DiffSplat~\cite{lin2025diffsplatrepurposingimagediffusion} adapts pre-trained text-to-image latent diffusion models to generate Structured Splats. Among 3D-native models, TRELLIS~\cite{xiang2024structured} generates 3DGS from a sparse voxel latent space, while GaussianCube~\cite{zhang2024gaussiancube} employs a 3D UNet to learn Gaussian distributions within a predefined voxel grid. We also compare against DreamGaussian~\cite{tang2023dreamgaussian}, an optimization-based framework that utilizes progressive densification, and LGM~\cite{tang2024lgm}, a feed-forward reconstruction model that we adapt for text-to-3D tasks by pairing it with a multi-view diffusion model~\cite{shi2023MVDream}.

\paragraph{Results and Comparisions:} As demonstrated in Tab.~\ref{tab:t3bench} and Fig.~\ref{fig:text_qualitative}, PixGS exhibits strong prompt alignment and visual quality among text-conditioned 3DGS generators. By inheriting 2D generative priors and bypassing latent compression such in DiffSplat, our approach successfully handles complex prompts. Conversely, 3D-native architectures are hindered by the limited availability of text-3D pairs for training from scratch. Both reconstruction and optimization-based frameworks suffer from multi-view inconsistencies due to their reliance on a 2D diffusion models, which struggle to maintain geometric coherence.
\begin{table}[t]
\centering
\caption{Quantitative evaluations on T3Bench. $\dagger$ denotes reconstruction-based methods requiring additional text-conditioned multi-view generative models. GC and DG denote GaussianCube and DreamGaussian, respectively. \textbf{Bold} indicates best, \underline{underline} second best.}
\label{tab:t3bench}
\footnotesize
\setlength{\tabcolsep}{2.5pt} 
\begin{tabular}{l c c c c cccc}
\toprule
Method & \multicolumn{2}{c}{DiffSplat~\cite{lin2025diffsplatrepurposingimagediffusion}} & \multicolumn{2}{c}{TRELLIS~\cite{xiang2024structured}} & GC & LGM$^\dagger$ & DG & PixGS\\ 
 & SD-1.5 & SD-3.5 & L & XL & \cite{zhang2024gaussiancube}& \cite{tang2024lgm}& \cite{tang2023dreamgaussian}& (Ours) \\ 
\midrule
\textit{Single Object} & & & & & & & & \\
$^\uparrow$CLIP Sim$_\%$   & 30.63 & \underline{30.99} & 28.77 & 28.75 & 27.35 & 29.96 & 24.78 & \textbf{31.98} \\
$^\uparrow$CLIP R-Pre$_\%$ & 78.50 & \underline{84.75} & 68.50 & 69.50 & 50.75 & 78.00 & 37.75 & \textbf{88.75} \\
$^\uparrow$ImgReward       & -0.490 & \underline{-0.196} & -1.033 & -0.961 & -1.521 & -0.720 & -1.635 & \textbf{-0.171} \\
\textit{Single w/ Surr} & & & & & & & & \\
$^\uparrow$CLIP Sim$_\%$   & 30.20 & \underline{30.91} & 28.37 & 28.35 & 25.60 & 27.79 & 23.81 & \textbf{31.32} \\
$^\uparrow$CLIP R-Pre$_\%$ & 86.00 & \textbf{87.75} & 73.50 & 70.00 & 47.25 & 55.00 & 41.50 & \underline{87.25} \\
$^\uparrow$ImgReward       & -1.063 & \underline{-0.447} & -1.604 & -1.518 & -2.063 & -1.772 & -1.953 & \textbf{-0.341} \\
\textit{Multiple Objects} & & & & & & & & \\
$^\uparrow$CLIP Sim$_\%$   & 27.62 & \underline{29.44} & 26.28 & 26.58 & 23.93 & 27.07 & 23.97 & \textbf{30.45} \\
$^\uparrow$CLIP R-Pre$_\%$ & 69.25 & \underline{74.00} & 48.00 & 49.75 & 26.25 & 51.00 & 28.50 & \textbf{77.25} \\
$^\uparrow$ImgReward       & -1.585 & \underline{-0.805} & -1.906 & -1.883 & -2.193 & -1.731 & -2.069 & \textbf{-0.740} \\\midrule
$^\downarrow$Latency (s)      & \textbf{1} & \underline{3} & 5 & 7 & \underline{3} & 6 & 47 & \textbf{1} \\ 
\bottomrule
\end{tabular}
\end{table}
\subsection{Image-conditioned Generation}
\begin{figure*}[]
    \centering
    \includegraphics[width=1.\textwidth]{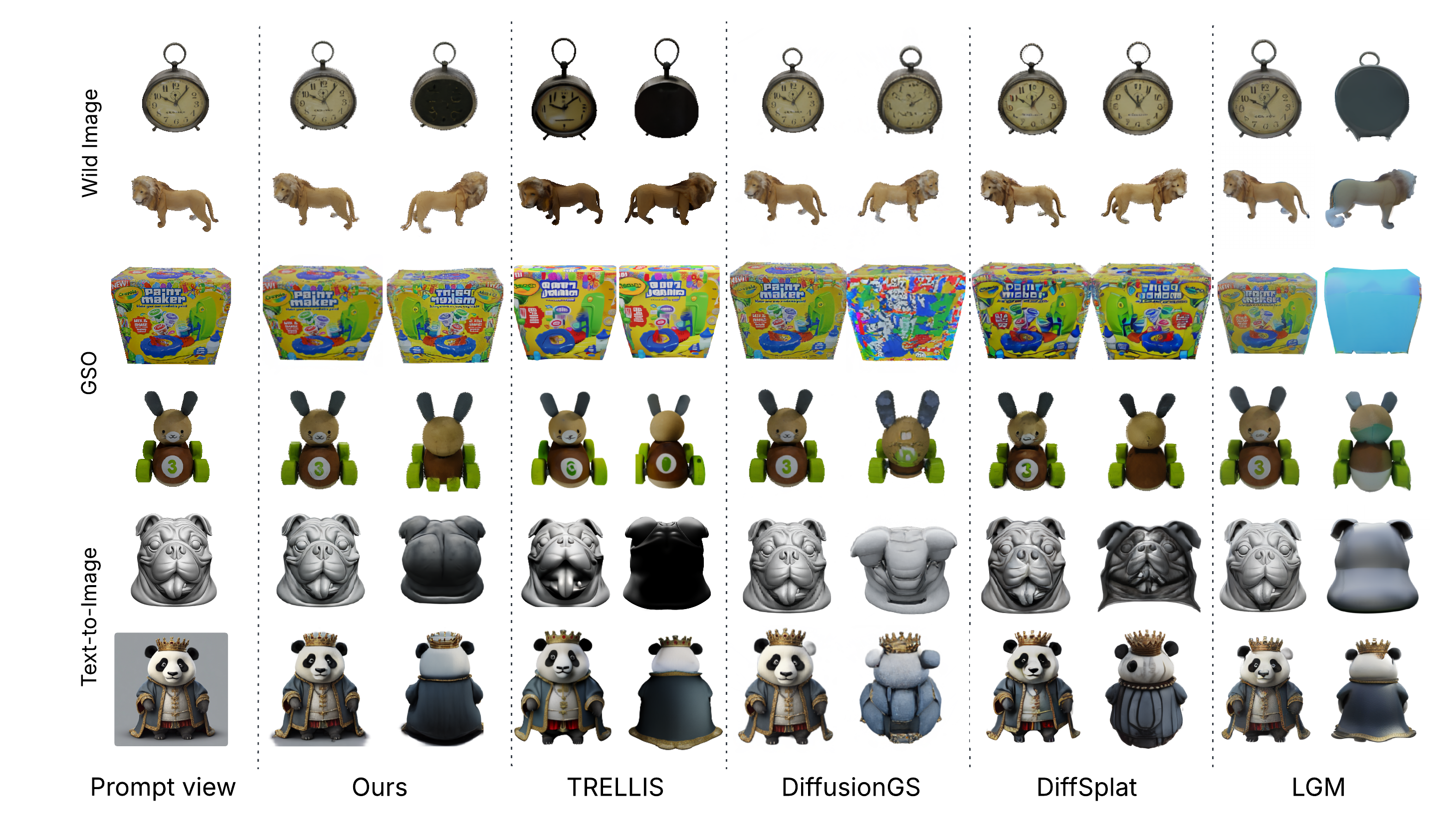}
    \caption{\textbf{Qualitative Results and Comparisons on Image-conditioned 3D Generation.} Best viewed \textbf{ZOOMED-IN}.}
    \label{fig:image_qualitative}
\end{figure*}
\paragraph{Baseline:} For the image-conditioned task, we compare our method against four representative baselines: DiffSplat~\cite{lin2025diffsplatrepurposingimagediffusion}, TRELLIS~\cite{xiang2024structured}, a single-stage pipeline DiffusionGS~\cite{diffusiongs}, and LGM~\cite{tang2024lgm} as a reconstruction-based method that relies on a separate multi-view diffusion model~\cite{shi2023MVDream} conditioned on an image prompt.

\paragraph{Results and Comparisions:} Quantitative and qualitative results (Tab.~\ref{tab:gso_metrics}, Fig.~\ref{fig:image_qualitative}) show that PixGS obtains favorable results in alignment and geometric fidelity. Unlike DiffusionGS, which struggles to generalize to unseen images due to being trained from scratch on limited corpus of synthetic 3D views. Furthermore, we observe that 3D-native models like TRELLIS face challenges in reference-view alignment. Because of operating in a canonical latent space without explicit viewpoint-conditional supervision, it often fails to reproduce the exact spatial orientation and perspective required to match the ground-truth reference image.

\begin{table}[t]
\centering
\caption{Quantitative evaluations on the GSO dataset. $\dagger$ denotes methods requiring additional image-conditioned multi-view generative models. \textbf{Bold} indicates best result, \underline{underline} second best.}
\label{tab:gso_metrics}
\footnotesize
\setlength{\tabcolsep}{4.5pt}
\begin{tabular}{l cccccc}
\toprule
Method & \multicolumn{2}{c}{DiffSplat~\cite{lin2025diffsplatrepurposingimagediffusion}} & DiffusionGS & TRELLIS-L & LGM$^\dagger$ & PixGS \\
& SD-1.5 & SD-3.5 &\cite{diffusiongs}& \cite{xiang2024structured}& \cite{tang2024lgm}& (Ours) \\ 

\midrule
$^\uparrow$PSNR           & 19.57 & \underline{19.59} & 16.53 & 17.70 & 15.61 & \textbf{21.21} \\
$^\uparrow$SSIM           & 0.810 & \underline{0.811} & 0.770 & 0.797 & 0.764 & \textbf{0.842} \\
$^\downarrow$LPIPS          & 0.159 & \underline{0.158} & 0.217 & 0.172 & 0.245 & \textbf{0.123} \\\midrule
$^\downarrow$Latency (s)    & \textbf{1} & 3 & 12 & 6 & \underline{2} & \textbf{1} \\
\bottomrule
\end{tabular}
\end{table}

\section{Ablation}
We analyze the efficacy of our image-conditioning strategies, loss functions, multi-phase training schedule and pseudo-label dependence through a series of ablation experiments. These models are trained on the G-Objaverse~\cite{qiu2023richdreamer, objaverse} dataset (265K assets) and evaluated on subset of 1,000 samples from G-Objaverse-XL Alignment~\cite{qiu2023richdreamer, objaverseXL}.
\subsection{Training Strategy}
\label{sec:ablation_loss}
\begin{figure*}[!h]
    \centering
    \includegraphics[width=1.\textwidth]{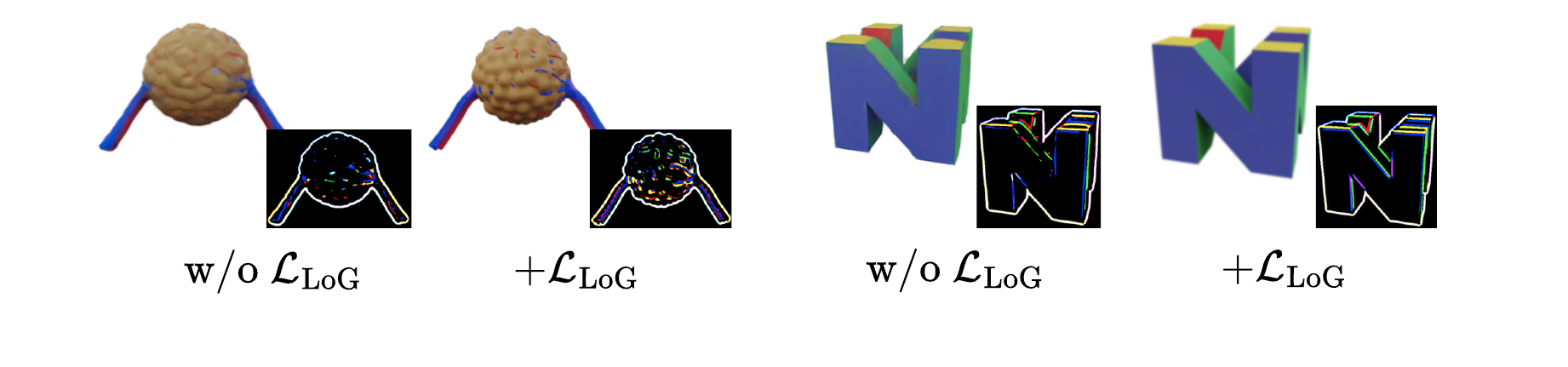}
    \caption{\textbf{Effect of $\mathcal{L}_{\text{LoG}}$.} $\mathcal{L}_{\text{LoG}}$ promotes the recovery of high-frequency details and mitigates over-smoothing, resulting in sharper geometric boundaries and texture.}
    \vspace{-20pt}
    \label{fig:ablation_logloss}
\end{figure*}

\begin{figure*}[t]
    \centering
    \includegraphics[width=0.8\textwidth]{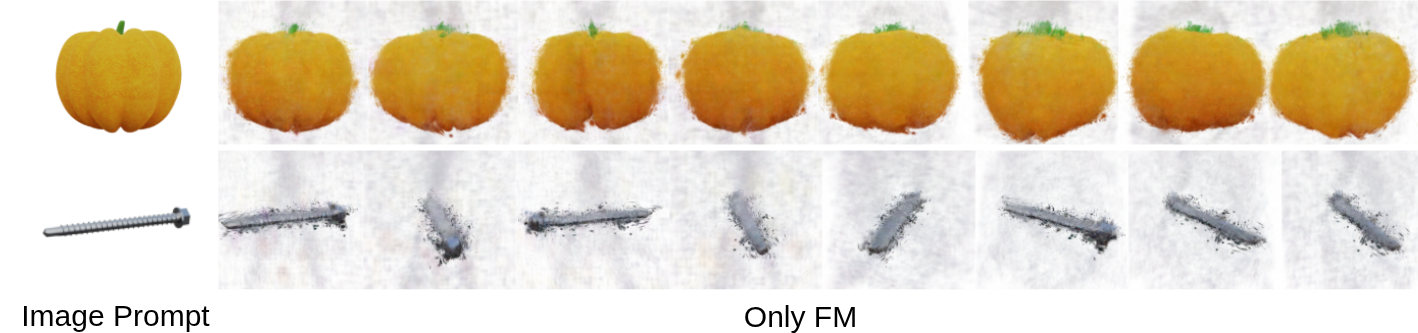}
    \caption{\textbf{Limitations of standalone Diffusion Loss supervision.} Despite the low loss values in Phase 1, the resulting 3DGS suffers from significant ``floater" artifacts.}    
    \label{fig:only_fm}

\end{figure*}

Tab.~\ref{tab:ablation_loss} highlights that adapting 2D pixel-space diffusion to the 3DGS modality requires more than naive distribution matching. Although optimizing only the flow-matching objective on pseudo-labels leads to rapid loss convergence, it does not guarantee high-quality 3D synthesis. Since our model operates directly at the splat level, small errors in attribute prediction, especially for near-zero opacity values, can produce persistent ``floater'' artifacts (see Fig.~\ref{fig:only_fm}). Thus, pseudo-label supervision is most useful as early guidance, while the main training signal comes from ground-truth appearance supervision. Geometry loss is auxiliary refinement for depth and normal quality, and the rotation regularizer stabilizes Gaussian parameterization, preventing numerical instability. Furthermore, the integration of the Multi-Scale LoG loss encourages the model to preserve high-frequency details, as shown in Fig.~\ref{fig:ablation_logloss}. Complementary to this, Tab.~\ref{tab:ablation_phases} shows that the three-phase training schedule gradually shifts the model from pseudo-label-guided pretraining to ground-truth-only supervision, thereby decoupling final generation quality from pseudo-label constraints.
\begin{table}[t]
\centering
\caption{\textbf{Ablation of image conditioning paradigms.} We compare Viewpoint Concatenation and Image Prompt Adaptation. \textbf{Bold} indicates best result, \underline{underline} second best.}
\label{tab:ablation_conditioning}
\footnotesize
\begin{tabular}{lcccc}
\toprule
Paradigm & $^\uparrow$PSNR  & $^\uparrow$SSIM  & $^\downarrow$LPIPS  & $^\downarrow$Params (B)  \\ \midrule
View Concatenation &  \textbf{30.36} & 0.969 & \textbf{0.022} & \textbf{1.2} \\ 
IP-Adapter &  30.31&  \textbf{0.977}& 0.023 & 1.4 \\ \bottomrule
\end{tabular}
\end{table}

\subsection{Image-conditioned Paradigms}
\label{sec:ablation_imagecond}
We systematically compare the two image-conditioning strategies in Tab.~\ref{tab:ablation_conditioning}. While both paradigms yield comparable performance, Viewpoint Concatenation is more parameter-efficient, facilitating training with larger batch sizes. We also find that Viewpoint Concatenation achieves higher PSNR, suggesting higher pixel-level fidelity. This is likely due to the preservation of the reference image's 2D spatial structure through view-axis concatenation, which enables denser and more direct cross-view attention. Conversely, Image Prompt Adapter approach demonstrates a slight advantage in capturing global structural patterns by leveraging the powerful priors of an image encoder.
\begin{table*}[!t]
    \centering
    \footnotesize
    \begin{minipage}{0.54\textwidth}
        \centering
        \caption{Ablation of supervision losses and regularization. \textbf{Bold} indicates best result.}
        \label{tab:ablation_loss}
        \begin{tabular}{lccc}
        \toprule
         & $^\uparrow$PSNR  & $^\uparrow$SSIM  & $^\downarrow$LPIPS  \\ \midrule
        Only $\mathcal{L}_{FM}$  & 16.04 & 0.708 & 0.496 \\ 
        + $\mathcal{L}_{app}$ & NaN & NaN & NaN \\ 
        + $\mathcal{L}_{rot}$ & 30.11 & \textbf{0.969} & 0.027 \\ 
        + $\mathcal{L}_{geo}$ & \underline{30.24} & \underline{0.967} & \underline{0.025} \\ 
        + $\mathcal{L}_{LoG}$ & \textbf{30.36} & \textbf{0.969} & \textbf{0.022} \\ \bottomrule
        \end{tabular}
    \end{minipage}
    \hfill
    \begin{minipage}{0.42\textwidth}
        \centering
        \caption{Effectiveness of the three-phase training curriculum. \textbf{Bold} indicates best result.}
        \label{tab:ablation_phases}
        \begin{tabular}{lccc}
        \toprule
         & $^\uparrow$PSNR  & $^\uparrow$SSIM  & $^\downarrow$LPIPS  \\ \midrule
        Phase 1 & 16.04 & 0.708 & 0.496 \\ 
        Phase 2 & \underline{30.36} & \underline{0.969} & \underline{0.022} \\ 
        Phase 3 & \textbf{31.15} & \textbf{0.987} & \textbf{0.019} \\ \toprule
        \end{tabular}
    \end{minipage}
\end{table*}
\subsection{Pseudo-label Dependence}
\begin{table}[!t]
\centering
\caption{Effect of pseudo-label source on PixGS's final quality on GSO. \textbf{Bold} indicates best result.}
\label{tab:ablation_pseudolabel}
\footnotesize
\begin{tabular}{lcccc}
\toprule
Pseudo-label & $^\uparrow$PSNR  & $^\uparrow$SSIM  & $^\downarrow$LPIPS  & $^\downarrow$Params (M)  \\ \midrule
LGM~\cite{tang2024lgm} &  21.15 & \textbf{0.844} & 0.125 & 415\\ 
GSRecon~\cite{lin2025diffsplatrepurposingimagediffusion} &  \textbf{21.21}&  0.842 & \textbf{0.123} &  \textbf{42}\\ \bottomrule
\end{tabular}
\end{table}
To assess PixGS's dependence on pseudo-labels from GSRecon~\cite{lin2025diffsplatrepurposingimagediffusion}, we replace them with lower-quality LGM~\cite{tang2024lgm} pseudo-labels. As shown in Tab.~\ref{tab:ablation_pseudolabel}, PixGS trained with LGM pseudo-labels achieves similar GSO performance, while requiring only about 10\% more training to reach plausible results. This suggests that our model's final quality is not strongly tied to the capability of the reconstructor. Therefore, we prefer GSRecon because it is compact and efficient, not because it sets the quality upper bound. Moreover, using pseudo-labels as early guidance avoids lengthy rendering-only optimization from noisy 3DGS point clouds with invalid parameters. Since higher-quality pseudo-labels can accelerate convergence, we incur a small, controlled GSRecon fine-tuning cost to improve training stability. This cost remains substantially lower than training a VAE required by existing latent-based methods.

\section{Conclusion}
We have presented PixGS, a single-stage model for 3DGS generation that bypasses the limitations of latent-space pipelines. By leveraging 2D pixel-space diffusion priors and a tailored suite of loss functions, including the multi-scale Laplacian of Gaussian loss, PixGS produces 3D objects with detailed appearance and strong geometric fidelity. Our three-phase training schedule and rotation regularization ensure stable convergence, enabling efficient generation of high-quality assets from complex text and image prompts in approximately one second. Overall, PixGS highlights the potential of direct splat-level denoising as a fast and effective approach for 3D generative modeling.

\par\vfill\par

\nocite{qwen2025qwen25technicalreport}
\nocite{ho2022classifierfreediffusionguidance}

\bibliographystyle{splncs04}
\bibliography{main}

\titlerunning{PixGS: Pixel-Space Diffusion for Direct 3D Gaussian Splat Generation}

\title{
PixGS: Pixel-Space Diffusion for Direct 3D Gaussian Splat Generation\\[2mm]
\textbf{\textit{(Supplementary Material)}}
}
\author{Cao Duy \qquad Phong Nguyen}
\authorrunning{C. Duy and P. Nguyen}
\renewcommand{\thefootnote}{\fnsymbol{footnote}}
\institute{
Qualcomm AI Research\footnotemark[1]\\
\texttt{\{duycao, phongnh\}@qti.qualcomm.com}
}
\footnotetext[1]{Qualcomm AI Research is an initiative of Qualcomm Technologies, Inc.}
\renewcommand{\thefootnote}{\arabic{footnote}}

\maketitle

\section{Implementation Details}
\paragraph{Architecture:} PixGS follows the backbone of PixNerd~\cite{wang2025pixnerdpixelneuralfield} with a patch size of $P=16$, using Qwen2.5-1.7B~\cite{qwen2025qwen25technicalreport} for text encoding and DINO-v2 ViT-B/14~\cite{oquab2023dinov2} for image encoding in image prompt adapter setting. We set number of viewpoints $V_{in} = 4$ with a $256 \times 256$ attribute grid, totaling 262,144 Gaussian splats per object. For multi-scale LoG loss, we select ${S} = \{1., 2., 3.\}$ with scale-specific weights $\omega_\sigma = \{0.05, 0.1, 0.15\}$.

\paragraph{Training:} We train PixGS using 8 H100 GPUs with an effective batch size of 128 using bf16 mixed precision. Training follows the three-phase schedule described in Sec. 4.4. Phase 1 converges rapidly, followed by refinement in Phases 2 and 3 with a total training duration of 5 days. We employ AdamW ($\beta{=}0.9, 0.999$, $\epsilon{=}10^{-8}$) with a peak learning rate of $10^{-5}$ and warmup for 1000 steps. Our model is initialized with pre-trained PixNerd~\cite{wang2025pixnerdpixelneuralfield} weights, specifically preserving the parameters for RGB channels and zero-initializing remaining channels corresponding to Gaussian attributes. For pseudo-label generation, we fine-tune GSRecon~\cite{lin2025diffsplatrepurposingimagediffusion} on the combined G-Objaverse~\cite{qiu2023richdreamer, objaverse} and G-Objaverse-XL~\cite{objaverseXL} Alignment datasets, encompassing over 1M objects.

\paragraph{Inference:} We use 25 denoising steps with classifier-free guidance~\cite{ho2022classifierfreediffusionguidance} scales of 4.0 for text and 2.0 for image conditioning. At full precision, PixGS generates a 3D object in $\sim1s$ on a single A100 GPU.

In both image-conditioned paradigms, when inference, we use a Plücker coordinates $\mathcal{P}(\theta_{cond}, \phi_{cond})$ corresponding to the elevation of the conditioned image $\theta_{cond}$ and a reference azimuth $\phi_{cond} = 0$.  For the noisy tensor $\mathcal{G}_{t}$, each viewpoint $v \in \{1, \dots, V_{in}\}$ is assigned an elevation $\theta_{v} = \theta_{cond}$, while $\phi_v$ are evenly distributed across the $360^\circ$ range:
\begin{equation}
    \phi_v = \frac{2\pi}{V_{in}}(v-1), \quad v = 1, \dots, V_{in}.
\end{equation}
We set first viewpoint’s azimuth to $\phi_1 = \phi_{cond}$ during both training and inference to ensure that the generated 3DGS is aligned with the coordinate system of the conditioned image.

\section{More Experiments}
\subsection{User study}
We conduct a user study to evaluate 3DGS generation methods based on human preference. The study considers two settings: text-conditioned generation and image-conditioned generation. Participants are presented with side-by-side $360^{\circ}$ turntable animations and asked to select the best 3D object based on each of three criteria: Prompt Alignment, Geometric Quality, and Textural Quality. Prompt Alignment evaluates the semantic and visual alignment of the generated asset to the input conditioning, ensuring that all descriptive details from text or key features from a reference image are preserved. Geometry and Texture Quality assess the structural and visual realism of the 3D model, rewarding smooth geometry and consistent textures. Specifically, we generated one sample for each text or image prompt, and the resulting outputs were included in the study without any manual selection or filtering. The study involved 103 participants, each completing 40 trials (20 per setting), resulting in a total of 4,120 independent evaluations. Detailed statistical results are provided in Table~\ref{tab:user_study_stats_text} and Table~\ref{tab:user_study_stats_image}.
\begin{table}[t]  
	\centering  
	\scriptsize
    \caption{Detailed statistics of the user study of Text-to-3D.}
    \vspace{-8pt}
	\setlength{\tabcolsep}{5pt}
	\begin{tabular}{l|cccc|c}  
		\toprule 
		\textbf{Method} & GaussianCube~\cite{zhang2024gaussiancube} & DiffSplat~\cite{lin2025diffsplatrepurposingimagediffusion} & TRELLIS~\cite{xiang2024structured} & Ours & Total \\
		\midrule
		\textit{Prompt Alignment} & & & & & \\
        $^\uparrow$Selections & 63 & \underline{527} & 121 & \textbf{1349} & 2060 \\
        $^\uparrow$Percentage$_\%$ & 3.1 & \underline{25.6} & 5.9 & \textbf{65.5} & 100 \\
		\midrule
		\textit{Geometric Quality} & & & & & \\
        $^\uparrow$Selections & 61 & 397 & \underline{413} & \textbf{1189} & 2060 \\
        $^\uparrow$Percentage$_\%$ & 3.0 & 19.3 & \underline{20.0} & \textbf{57.7} & 100 \\
        \midrule
		\textit{Textural Quality} & & & & & \\
        $^\uparrow$Selections & 67 & \underline{487} & 143 & \textbf{1363} & 2060 \\
        $^\uparrow$Percentage$_\%$ & 3.3 & \underline{23.6} & 7.0 & \textbf{66.2} & 100 \\
		\bottomrule
	\end{tabular}  
	\label{tab:user_study_stats_text}  
\end{table}
\begin{table}[t]  
	\centering  
	\scriptsize
    \caption{Detailed statistics of the user study of Image-to-3D.}
    \vspace{-8pt}
	\setlength{\tabcolsep}{5pt}
	\begin{tabular}{l|cccc|c}  
		\toprule 
		\textbf{Method} &  DiffusionGS~\cite{diffusiongs}& DiffSplat~\cite{lin2025diffsplatrepurposingimagediffusion} & TRELLIS~\cite{xiang2024structured} & Ours & Total \\
		\midrule
		\textit{Prompt Alignment} & & & & & \\
        $^\uparrow$Selections  & 27 & 83 & \underline{613} & \textbf{1337} & 2060 \\
        $^\uparrow$Percentage$_\%$ & 1.3 & 4.0 & \underline{29.8} & \textbf{65.0} & 100 \\
		\midrule
		\textit{Geometric Quality} & & & & & \\
        $^\uparrow$Selections & 43 & 109 & \textbf{981} & \underline{927} & 2060 \\
        $^\uparrow$Percentage$_\%$ & 2.1 & 5.3 & \textbf{47.6} & \underline{45.0} & 100 \\
        \midrule
		\textit{Textural Quality} & & & & & \\
        $^\uparrow$Selections & 89 & 211 & \underline{863} & \textbf{897} & 2060 \\
        $^\uparrow$Percentage$_\%$ & 4.3 & 10.2 & \underline{41.9} & \textbf{43.5} & 100 \\
	\bottomrule
	\end{tabular}  
	\vspace{-8pt}
	\label{tab:user_study_stats_image}  
\end{table}

\subsection{Data Scalability}
\begin{table}[t]
\centering
\caption{\textbf{Scalability and architectural comparison.} We compare our method against baselines using the same set of 300 randomly selected samples from GSO dataset under image-conditioned settings.}
\label{tab:scalability}
\vspace{-8pt}
\setlength{\tabcolsep}{4pt}
\begin{tabular}{l ccc c cc}
\hline
Data Scale & \multicolumn{3}{c}{$\sim 265K$ samples} & & \multicolumn{2}{c}{$\sim 1M$ samples} \\
\cline{2-4} \cline{6-7}
Method & DiffSplat~\cite{lin2025diffsplatrepurposingimagediffusion} & DiffusionGS~\cite{diffusiongs} & Ours & & DiffusionGS ~\cite{diffusiongs}& Ours \\
\hline
$^\uparrow$PSNR & 19.59 & 15.28 & \textbf{20.67} & & 16.53 & \textbf{21.13} \\
$^\uparrow$SSIM & 0.811 & 0.633 & \textbf{0.827} & & 0.770 & \textbf{0.838} \\
$^\downarrow$LPIPS & 0.158 & 0.341 & \textbf{0.144} & & 0.217 & \textbf{0.130} \\
\hline
Single-stage     & --         & \checkmark & \textbf{\checkmark} & & \checkmark & \textbf{\checkmark} \\
w/ 2D Prior      & \checkmark & --         & \textbf{\checkmark} & & --         & \textbf{\checkmark} \\
\hline
\end{tabular}
\end{table}
Table~\ref{tab:scalability} demonstrates the impact of training data scale, comparing G-Objaverse~\cite{qiu2023richdreamer,objaverse} ($\sim 265K$ assets) with an extended dataset ($\sim 1M$ assets) that includes G-Objaverse-XL~\cite{objaverseXL} on the generative performance of our model relative to DiffSplat~\cite{lin2025diffsplatrepurposingimagediffusion} and DiffusionGS~\cite{diffusiongs}. These two baselines share a similar objective of generating 3DGS and are trained on multi-view rendered images. For this evaluation, we use the same 300 randomly selected samples from GSO dataset (Sec. 5.3) under image-conditioned settings.

Two-stage models such as DiffSplat, which require sequential training of an autoencoder and a diffusion model resulting in scalability limits imposed by expensive computation and engineering overhead. In those frameworks, the autoencoder must converge sufficiently well before diffusion training, creating a significant bottleneck. Conversely, single-stage specialized pipelines that train from scratch without 2D priors, such as DiffusionGS, suffer from slow convergence (requiring 32 A100 GPUs for 140K iterations~\cite{diffusiongs}) and exhibit poor performance when trained on smaller datasets.
Addressing both issues, PixGS is a single-stage pipeline that leverages large-scale 2D priors. This allows for rapid convergence without the need for separate module training, thereby facilitating large-scale scalability. Our model outperforms the baselines across both data scale settings, demonstrating the clear advantage of our proposed method.

\subsection{Sampling Diversity}
\begin{figure}[t]
    \centering
    \includegraphics[width=\textwidth]{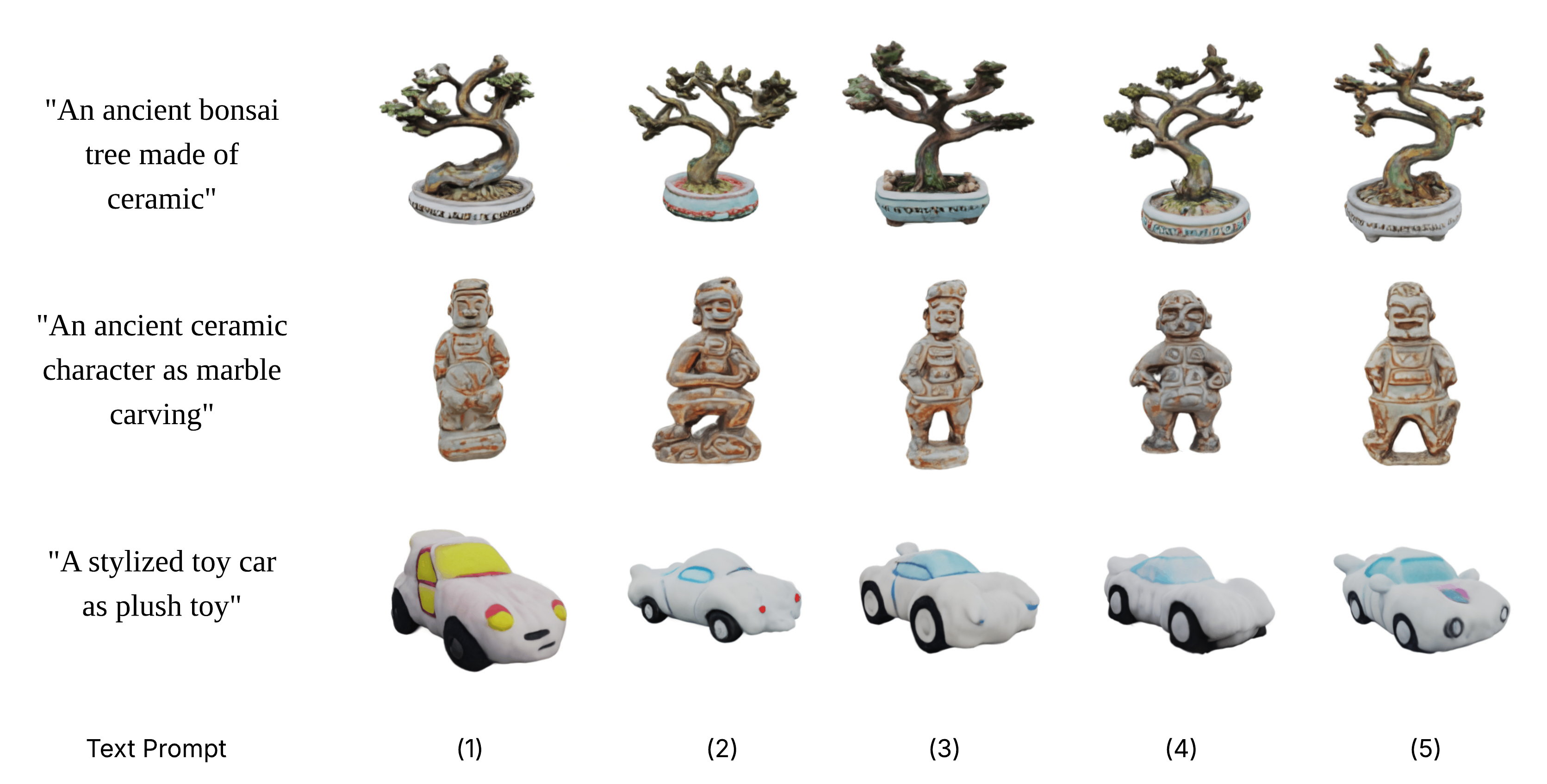}
    \caption{Generation results across different seeds.}
    \vspace{-10pt}
    \label{fig:diversity}
\end{figure}
We demonstrate the sampling diversity of PixGS in Fig.~\ref{fig:diversity}. As a single-stage generative model that learning directly at the splat level without an intermediate latent space, our model effectively captures the stochastic nature of the 3DGS distribution. When conditioned on the same text prompt with varying random seeds, PixGS synthesizes a diverse range of geometric and textural interpretations while consistently maintaining alignment with the provided conditioning.

\subsection{$\sigma$ choice for LoG loss}
\label{sec:ablation_logsima}
\begin{figure*}[]
    \centering
    \includegraphics[width=\textwidth]{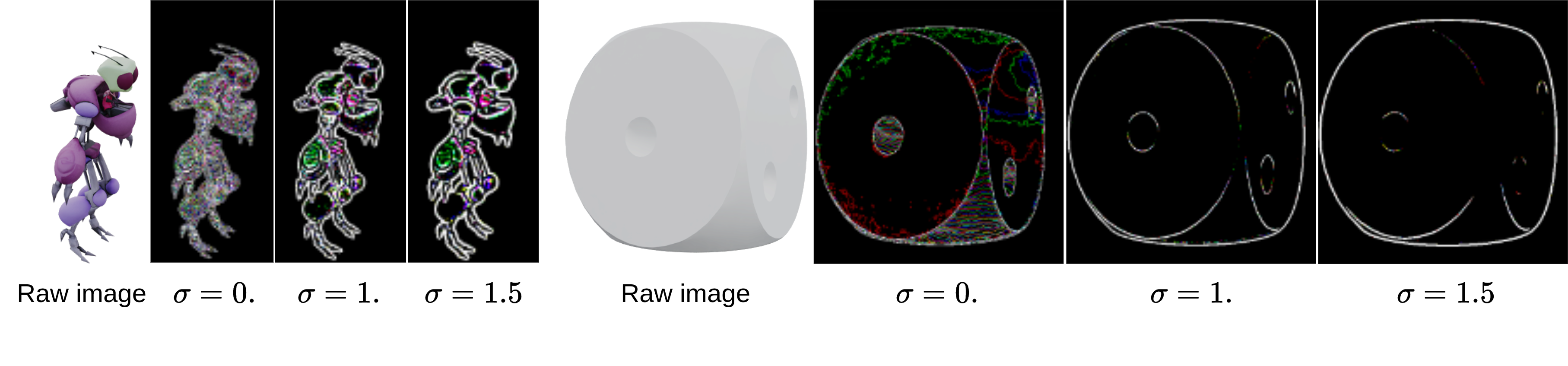}
    \vspace{-18pt}
    \caption{\textbf{Laplacian of Gaussian (LoG) feature extraction at multiple scales.} Each panel illustrates the filtered results across different standard deviations ($\sigma \in \{0, 1, 1.5\}$).}
    \label{fig:log_sigma}
\vspace{-10pt}
\end{figure*}

Fig.~\ref{fig:log_sigma} shows that smaller $\sigma$ kernels facilitate the recovery of sharp structural transitions but are susceptible to high-frequency artifacts. In contrast, larger $\sigma$ values prefer global object silhouette, but lack fine-grained textural components. We observe that applying LoG objective with small $\sigma$ values can lead the model to replicate stochastic rendering noise artifacts, therefore we prioritize stable gradients of larger $\sigma$ scales to supervise global structural coherence, while using a smaller weight for fine-scale kernels to resolve intricate features without amplifying high-frequency rendering noise.

\section{Limitations and Future Work}
\vspace{-15pt}
While our model achieves strong performance on 3D generation with rapid inference, several limitations remain. First, it is currently restricted to object-level generation. Given that representing scenes via 3D Gaussian Splats is crucial for many applications, extending this approach may necessitate increased supervision resolution or a higher number of viewpoints. Second, our approach does not explicitly model physically-based material properties, nor does our image-conditioned paradigm decouple environmental lighting. This results in "baked-in" shading and specular highlights derived from the reference image. Lastly, while our method's scalability is currently evaluated on synthetic multi-view datasets, it could be further enhanced by leveraging real-world video datasets or camera-controlled generative models, which we leave for future exploration.
\vspace{-20pt}
\section{More Qualitative Results}
\pagebreak
\begin{figure}[H]
    \centering
    \includegraphics[width=\textwidth]{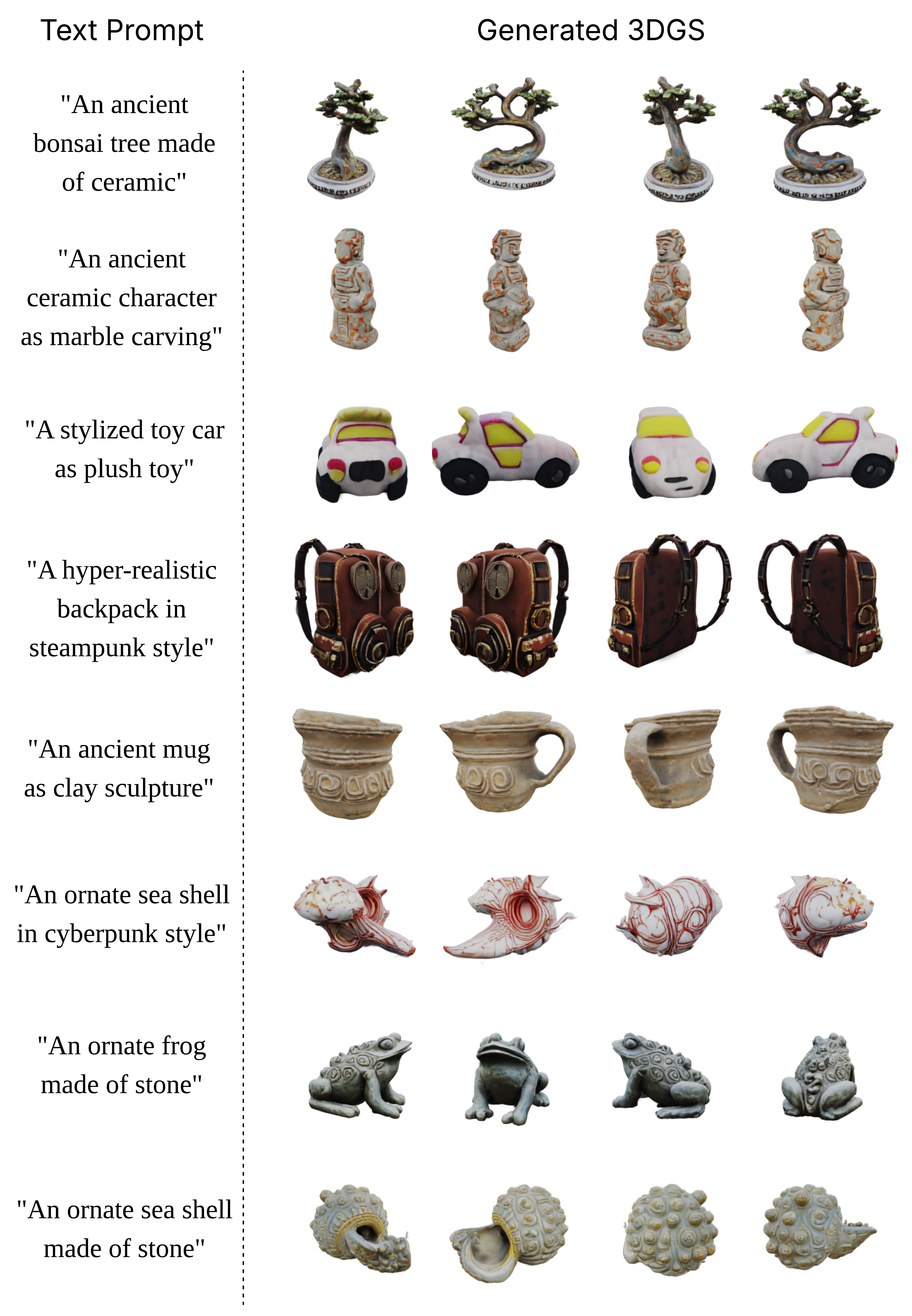}
    \caption{More text-conditioned results of PixGS}
    \label{fig:text_qualitative_1}
\end{figure}
\begin{figure}[H]
    \centering
    \includegraphics[width=\textwidth]{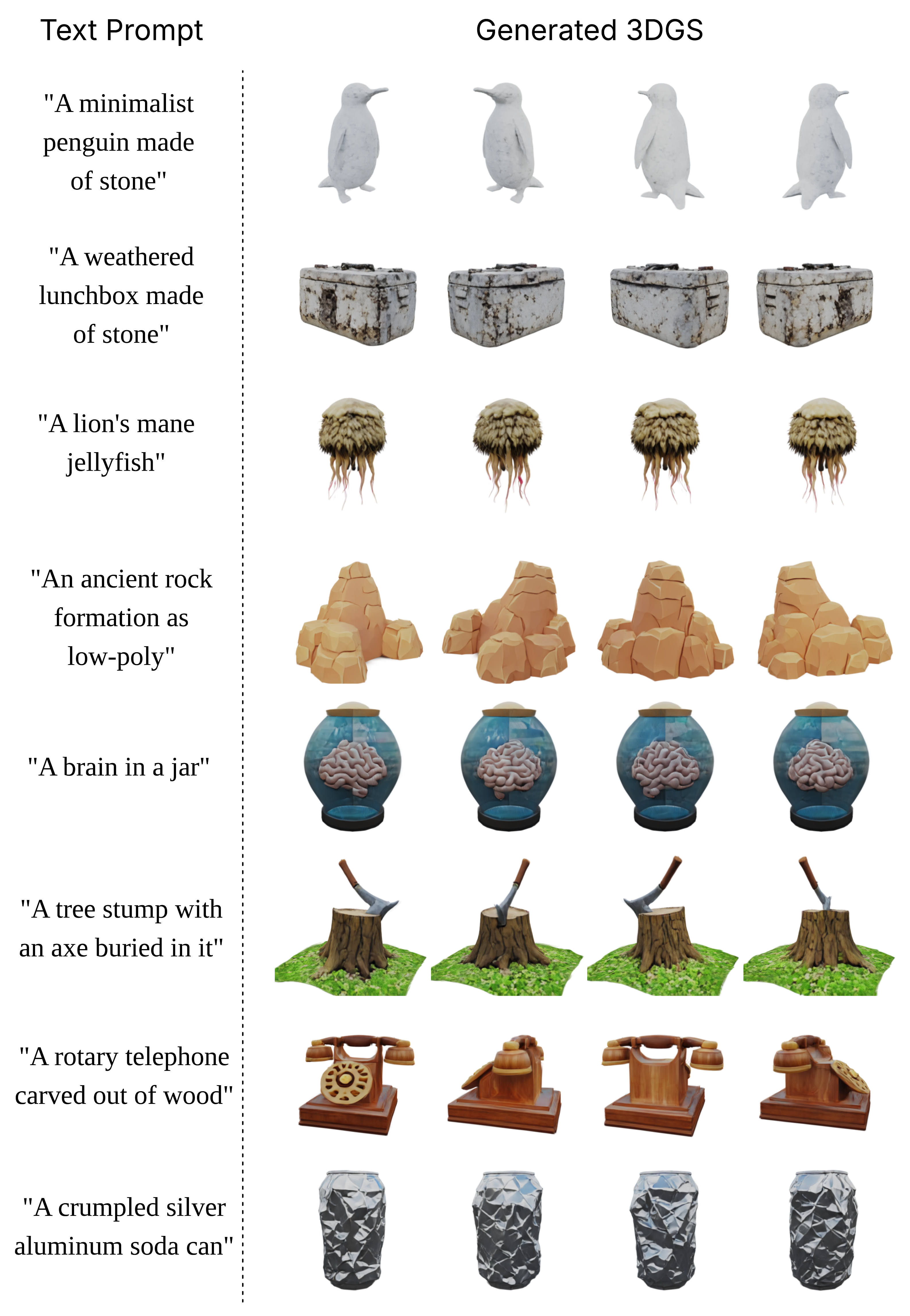}
    \caption{More text-conditioned results of PixGS}
    \label{fig:text_qualitative_2}
\end{figure}
\begin{figure}[H]
    \centering
    \includegraphics[width=\textwidth]{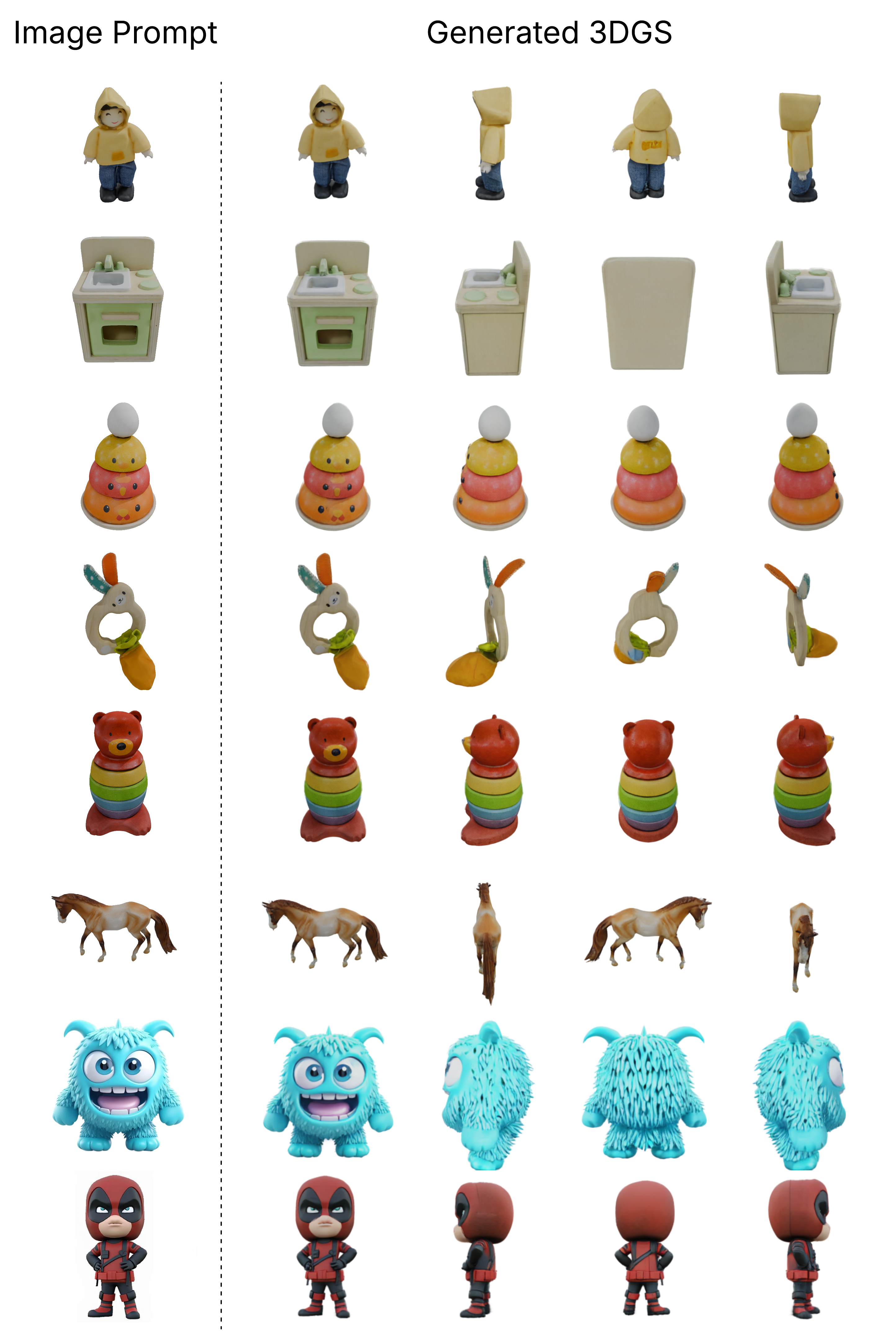}
    \caption{More image-conditioned results of PixGS}
    \label{fig:image_qualitative_1}
\end{figure}

\begin{figure}[H]
    \centering
    \includegraphics[width=\textwidth]{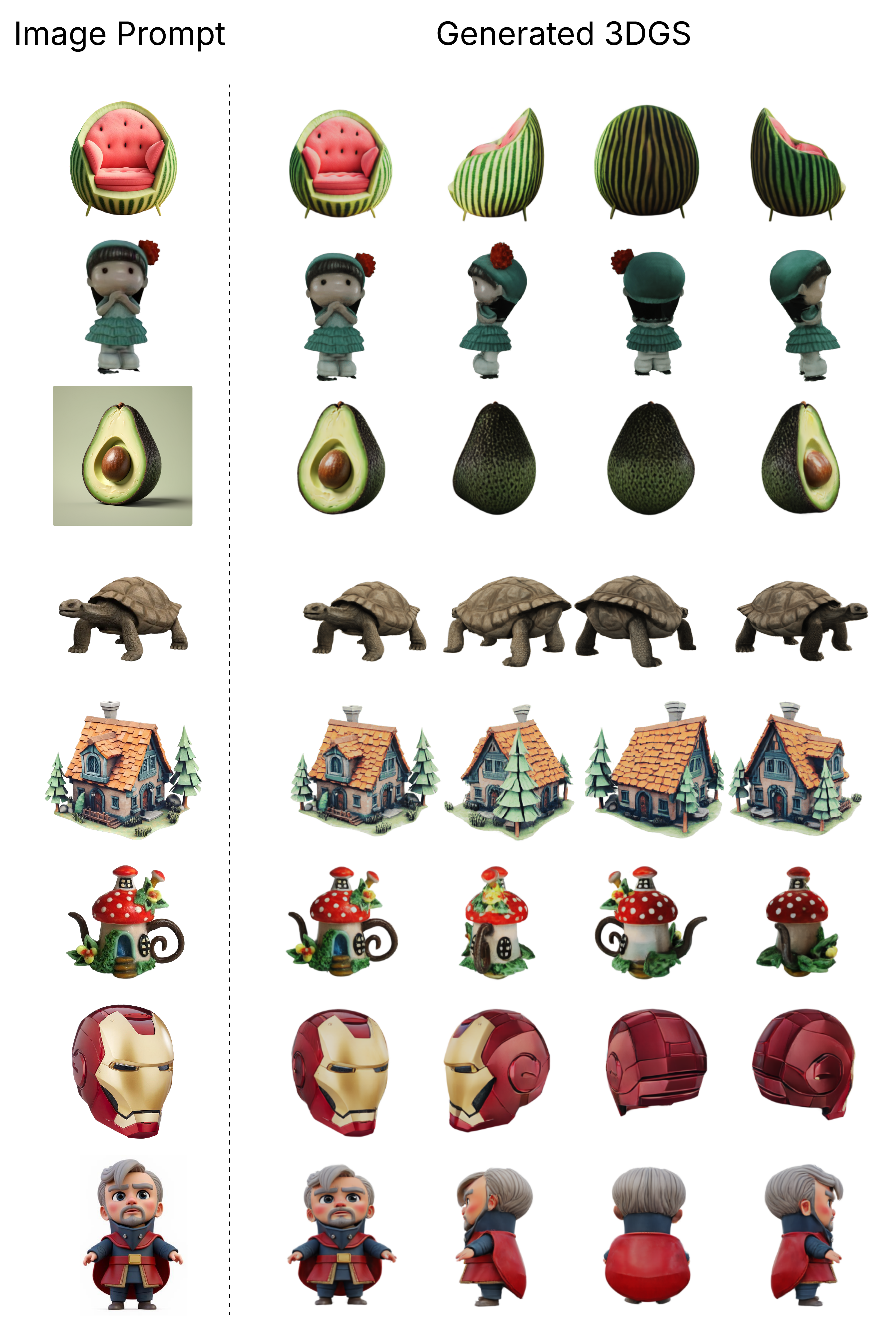}
    \caption{More image-conditioned results of PixGS}
    \label{fig:image_qualitative_2}
\end{figure}


\end{document}